\documentclass{article}

\usepackage[preprint]{neurips_2023}

\usepackage[utf8]{inputenc} % allow utf-8 input
\usepackage[T1]{fontenc}    % use 8-bit T1 fonts
\usepackage{hyperref}       % hyperlinks
\usepackage{url}            % simple URL typesetting
\usepackage{booktabs}       % professional-quality tables
\usepackage{amsfonts}       % blackboard math symbols
\usepackage{nicefrac}       % compact symbols for 1/2, etc.
\usepackage{microtype}      % microtypography
\usepackage{xcolor}         % colors
\usepackage{graphicx}
\usepackage{amsmath}
\usepackage{multirow}

\title{GraphKAN: Enhancing Feature Extraction with Graph Kolmogorov Arnold Networks}

\author{%
  Fan Zhang\\
  Department of Mathematics\\
  The Hong Kong University of Science and Technology\\
  \texttt{mafzhang@ust.hk} \\
  \AND
  Xin Zhang* \\
  Department of Mechanical and Aerospace Engineering\\
  The Hong Kong University of Science and Technology\\
  \texttt{mexzyl@ust.hk} \\
}

\begin{document}

\maketitle

\begin{abstract}
  Massive number of applications involve data with underlying relationships embedded in non-Euclidean space. Graph neural networks (GNNs) are utilized to extract features by capturing the dependencies within graphs. Despite ground-breaking performances, we argue that Multi-layer perceptrons (MLPs) and fixed activation functions impede the feature extraction due to information loss. Inspired by Kolmogorov Arnold Networks (KANs), we make the first attempt to GNNs with KANs. We discard MLPs and activation functions, and instead used KANs for feature extraction. Experiments demonstrate the effectiveness of GraphKAN, emphasizing the potential of KANs as a powerful tool. Code is available at \url{https://github.com/Ryanfzhang/GraphKan}.
\end{abstract}

\section{Introduction}

The recent success of neural networks \cite{o2015introduction,deng2009imagenet} has greatly stimulated research in the fields of pattern recognition \cite{wang2017residual,stahlberg2020neural} and data mining\cite{he2017neural}. Especially, in many applications \cite{cui2019traffic,fout2017protein}, data are generated from non-Euclidean domains. Complicated relationships between objects are represented as graphs. For example, in biology, protein can be viewed  as graphs \cite{gligorijevic2021structure}, and their structure needs to be identified for protein-protein interaction; in social networks \cite{fan2019graph}, interactions between peoples indicate similar interests or attributes.

Unfortunately, complexity of graph data has presented significant challenges for traditional deep learning algorithms \cite{zhou2020graph}. Nodes in an irregular graph have varying numbers of neighbors. Handing complex patterns of neighbors and extracting feature of each node are troublesome \cite{xu2018powerful}. Consequently, there has been considerable scholarly interest in harnessing deep learning techniques for the purpose of feature extraction from intricate graph structures.

As an excellent feature extractor for graph, graph neural networks (GNNs) \cite{scarselli2008graph} are capable of handing the characteristics of the graph due to the message passing framework \cite{vignac2020building}. Concretely, message passing framework for node representation iteratively (1) \textbf{aggregates} information from the neighbors and (2) \textbf{extracts} the representations of nodes. Previous studies have made significant efforts in aggregating information, such as GCN \cite{zhang2019graph}, GAT \cite{velivckovic2017graph} and so on. Diverse aggregation approaches facilitate the incorporation of multiple features, including node importance and heterogeneity \cite{zhang2019heterogeneous}, thereby leading to enhanced efficiency in information passing.

Despite advancements, there are still inherent limitations in the process of extracting representations (Step 2 in message passing framework). In general, the majority of methods utilize Multi-layer Perceptron (MLP) \cite{taud2018multilayer} for feature extraction. To enhance the nonlinearity capability, an activation function is commonly incorporated between two graph convolutional layer. Regrettably, MLP in the majority of the aforementioned models suffer from poor scaling laws \cite{friedland2017capacity}. Specifically, the number of parameters in MLP networks does not scale linearly with the number of layers, leading to a lack of interpretability \cite{barcelo2020model}. Moreover, activation function, such as ReLU, limit the representational capacity \cite{yarotsky2017error}, potentially preventing it from learning complex features of nodes. To conclude, MLPs+activation function impede the feature extraction in graph-like data.

A recent study introduces Kolmogorov-Arnold Networks (KANs) \cite{liu2024kan}, a novel neural network architecture designed to potentially replace traditional MLPs. Unlike MLPs, KANs introduce a novel approach by substituting linear weights with spline-based univariate functions along the edges of the network. These functions are specifically structured as learnable activation functions. KANs offer a solution to the limitations of efficiency and interpretability encountered in MLPs in GNNs. Additionally, KANs address the problem of information loss resulting from the use of activation functions. 

In this paper, we firstly introduce KANs to feature extraction in graph-like data, dubbed GraphKAN. We replace all MLPs and activation function with KANs in GNNs. Moreover, we add LayerNorm \cite{xu2019understanding} to stabilize the learning process. To evaluate the practicality of GraphKAN in real- world scenarios, we assess the performance using real-world graph-like temporal signal data for signal classification. Experiments demonstrate the effectiveness of GranKAN in feature extraction.

\section{Problem Statement}
To evaluate the practicality of GraphKAN in real-world scenarios, we choose a signal classification task. 

For a dataset $D = {(X_1, Y_1), (X_2, Y_2), \cdots, (X_n, Y_n)}$, $X$ and $Y$ represent the raw one-dimension (1-D) sampling signals and fault categories, respectively. We transform 1-D signals into the form of graph. The detailed construction process can refer as . And the constructed graph is named as basic graph (BG), which can be defined as \cite{zhang2024pruned,zhang2024multiscale}, where V represents the nodes of G, E is the edge connections between nodes, A indicates the adjacency matrix, and F means the feature matrix of G. As a result, our goal is now focused on recognizing node labels. We can evaluate the model's performance based on diagnosis accuracy. It is important to highlight that the node classification task we have adopted is not tailored to a specific design, and the proposed GraphKAN can enhance feature extraction capabilities as a versatile tool.

\section{Methodology}
In this section, we detail the overall architecture of our GraphKAN. Kolmogorov-Arnold Network provides a powerful information aggregation in Graph Neural Network. Due to the enhanced representation ability, GraphKAN is more suitable for Graph-like tasks, such as node classification, graph classification and so on.  

\subsection{Kolmogorov-Arnold Network (KAN)}
Contrary to universal approximation theorem, Kolmogorov-Arnold representation theorem demonstrates that any multivariate continuous function $f$ on a bounded domain can be represented as the finite composition of simpler 1-D continuous functions:
\begin{equation}
	f(x_1,\cdots,x_n)=\sum_{q=1}^{2n+1}\Phi_q (\sum_{p=1}^n \phi_{q,p}(x_p)),
\end{equation}
where $\phi_{q,p}$ is a mapping $[0,1]\rightarrow \mathbb{R}$ and $\Phi_{q}$ is a mapping $\mathbb{R}\rightarrow \mathbb{R}$. However, these one-dimensional functions can exhibit non-smoothness and even fractal characteristics, making them impractical to learn in practice.

\cite{liu2024kan} devise a neural network that explicitly incorporates this equation.  This network is designed in such a way that all learnable functions are univariate and parameterized as B-splines, thereby enhancing the flexibility and learnability of the model.

Specifically, the Kolmogorov-Arnold Network (KAN) can be given by
\begin{equation}
	x_{l+1}=\Phi_l x_l,
\end{equation}
where $\Phi_l$ is a matrix
\begin{equation}
	\begin{pmatrix}
	\phi_{1,1} & \phi_{1,2} & \cdots & \phi_{1,n}\\
	\phi_{2,1} & \phi_{2,2} & \cdots & \phi_{2,n}\\
	\vdots & \vdots & \ddots & \vdots \\
	\phi_{m,1} & \phi_{m,2} & \cdots & \phi_{m,n}
\end{pmatrix}.
\end{equation}

\subsection{GraphKAN Layer}
Graph neural networks (GNNs) are capable of handing the characteristics of the graph due to the message passing framework. GNNs can process the graph-like data which contains of node features $x_v$ and edge features $e_{vw}$. In general, node features denote the attributes of node and edge features represent the intra-node relations in the form of adjacency matrix.

Message passing framework for node representation can be divided into two steps:
\begin{itemize}
	\item [1)] \textbf{Aggregates} information from the neighbors. The message function is used to aggregate the neighboring features  of the target node, including the target node's own features  $h_v^t$, the features of its neighboring nodes $h_w^t$, and the edge features connecting it to its neighboring nodes $e_{vw}^t$. This aggregation forms a message vector $m_v^t$ that is then passed to the target node. The formula is as follows:
\begin{equation}
	m_v^{t+1} = \sum_{w\in N(v)}M^t(h_v^t, h_w^t, e_{vw}^t)
\end{equation}
where $m_v^{t+1}$ is the information received by the node in the next layer $t+1$, $M^t$ is the message function, $h_v^t$ represents the node features in the current layer, $N(v)$ represents the set of neighboring nodes for a node $v$, $h_w^t$ represents the node features of the neighboring nodes in the current layer, and $e_{vw}^t$ represents the edge features from node to node.
	\item [2)] \textbf{Extracts} the node representation. The node update function is used to update the node features of the next layer, combining the features of the current layer's nodes and the messages obtained from the message function. The formula is as follows:
\begin{equation}
	h_v^{t+1} = U^t(h_v^t, m_v^{t+1})
\end{equation}
where $U^t$ is the node update function, which takes the original node state and the message as inputs and generates the new node state.
\end{itemize}

Note that, the majority of methods utilize Multi-layer Perceptron (MLP) for feature extraction. To enhance the nonlinearity capability, an activation function is commonly incorporated in the Phase 2) above. Activation function, such as ReLU, limit the representational capacity, potentially preventing it from learning complex features of nodes. To conclude, MLPs+activation function impede the feature extraction in graph-like data.

Hence, we propose to improve the $Extracts$ the node representation phase by replacing MLP with KAN for $U^t$. Our new representation extraction can be formalized as:
\begin{equation}
	h_v^{t+1} = KAN^t(h_v^t, m_v^{t+1})=\Phi^t(h_v^t, m_v^{t+1}).
\end{equation}

Moreover, we utilize the B-spline function as the $\Phi^t$. Despite many variants of KAN, such as FourierKAN and ChebyKAN, are proposed,  we found that they are not effective for graph tasks.

\subsection{A Simple GraphKAN for Node Classification}
To evaluate the ability of representation extraction of GraphKAN, we contrive a simple GraphKAN based on Graph Convolutional Neural Network (GCN).

Concretely, the formula of GraphKAN layer based on GCN is as follows:
\begin{itemize}
	\item [1)] message function $M^t$ is
\begin{equation}
	(deg(v) deg(w))^{-1/2} A_{uw}h_w^t,
\end{equation}
where $deg$ computes the degree of node.
    \item [2)] node representation extraction function $U^t$ is 
 \begin{equation}
 	U^t(h_v^t, m_v^{t+1})=\Phi^t m_v^{t+1}
 \end{equation}.
\end{itemize}

To stabilize the training, we leverage the LayerNorm between GraphKAN layers.

\section{Experimental Verification}
\label{Experiment}

\begin{table}
  \caption{Settings of four input graphs.}
  \label{tab:settings}
  \centering
  \begin{tabular}{cccc}
    \toprule
    &\multicolumn{2}{c}{Labeled node amount}                   \\
    \cmidrule(r){2-3}
    Graph     & Label 0     & Label 1-5 & Unlabeled Amount\\
    \midrule
    BG\_1 & 200  &  100  & 700 \\
    BG\_2     & 200 &   80&  700  \\
    BG\_3     & 200        & 60 & 700\\
    BG\_4     & 200       & 40    & 700\\
    \bottomrule
  \end{tabular}
\end{table}

\begin{table}
  \caption{Model structural settings.}
  \label{tab:model}
  \centering
  \begin{tabular}{ccc}
    \toprule
   Name & Model structural settings & Kernel size\\
   \midrule
   GCN & \multirow{2}{*}{GCN structure: 512-256-128, layer normalization} & \multirow{2}{*}{$k_1=3$, $k_1=3$,$k_1=3$}\\
   GraphKAN& \\
    \bottomrule
  \end{tabular}
\end{table}

\begin{figure}[!t]
\centering
\includegraphics[scale=1]{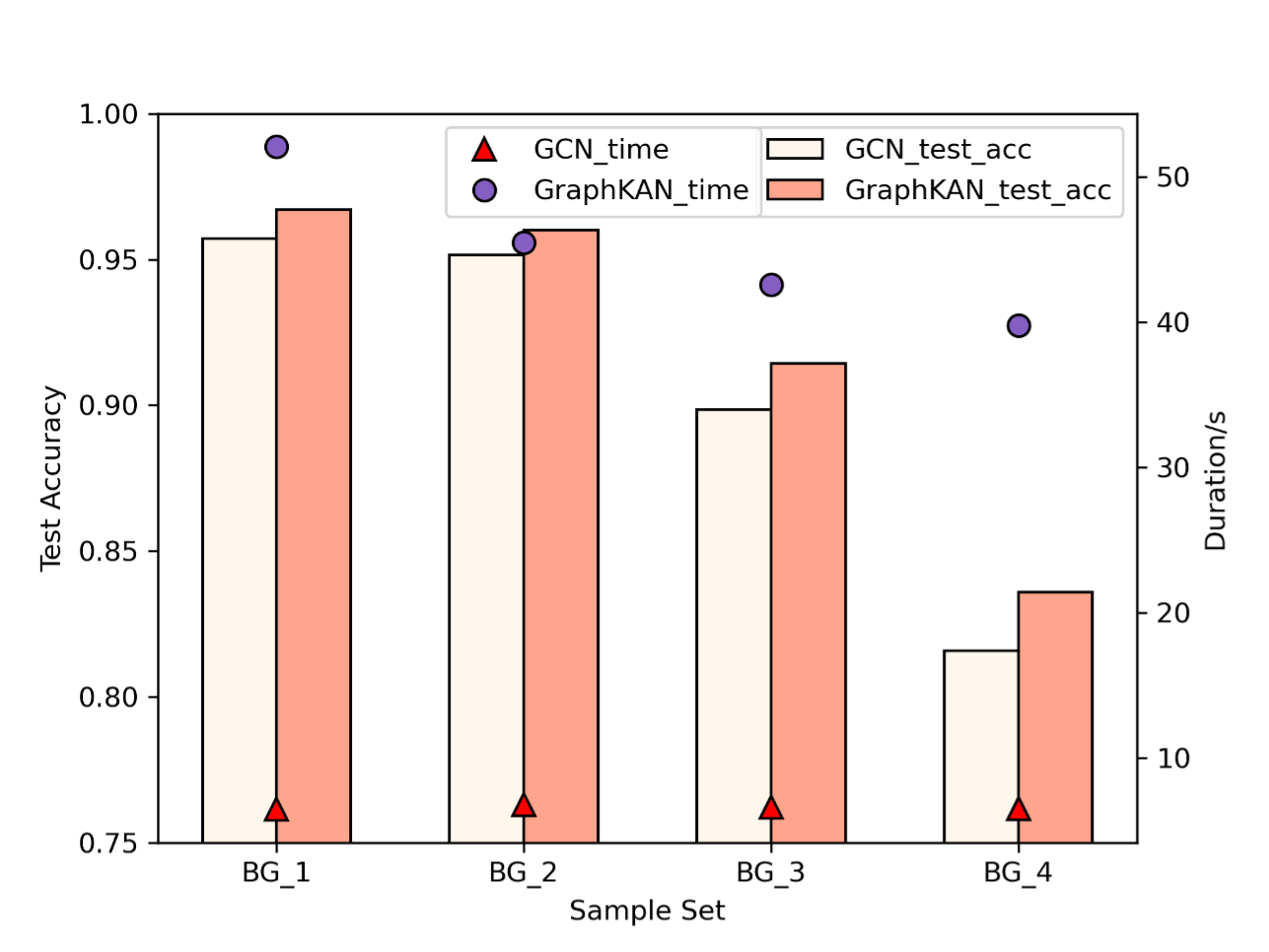}
\caption{Comparison on testing accuracy and time consumption}
\label{fig1}
\end{figure}

\begin{figure}[!t]
\centering
\includegraphics[scale=0.8]{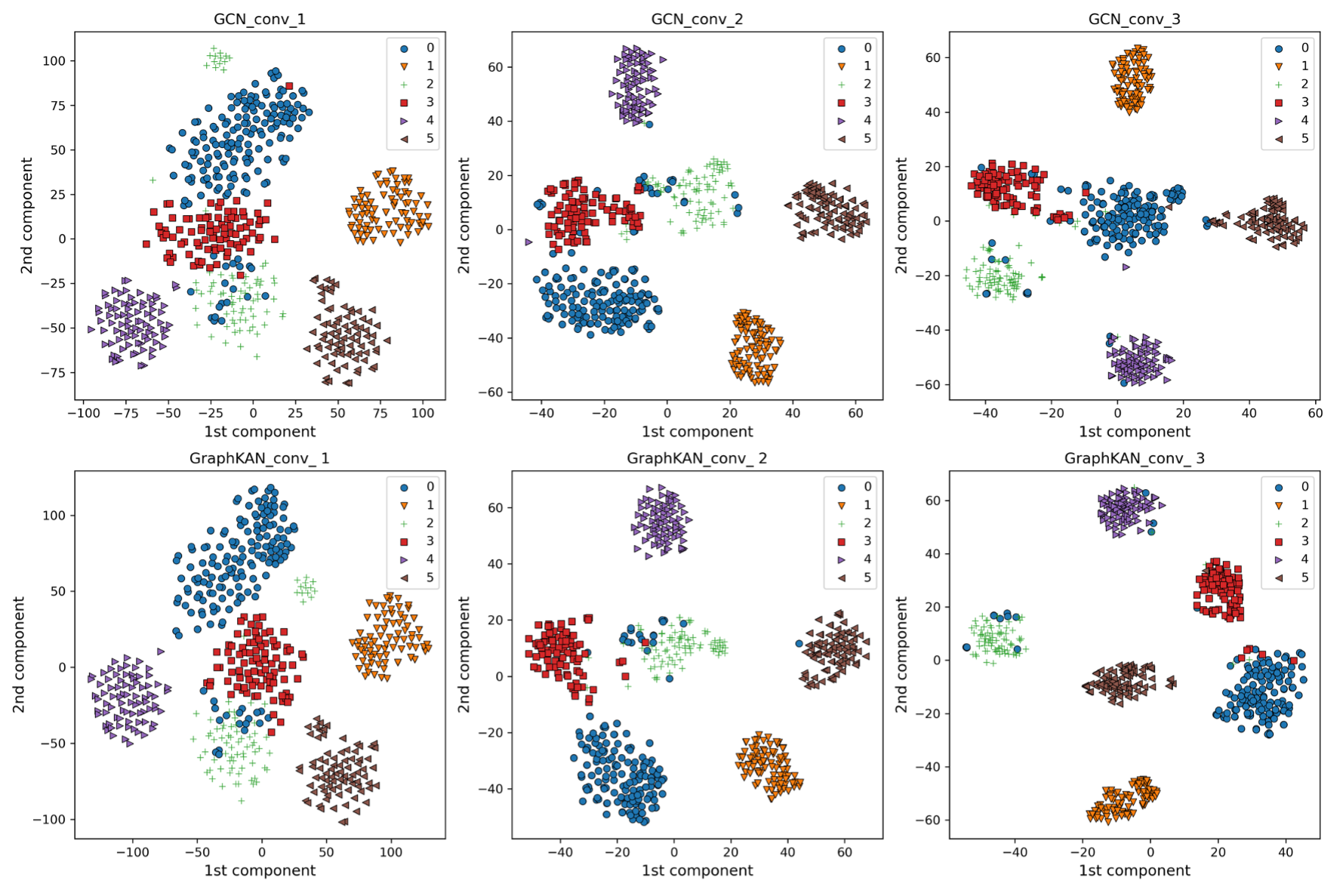}
\caption{Clustering comparison of intermediate features for testing nodes within BG\_1}
\label{fig2}
\end{figure}

\begin{figure}[!t]
\centering
\includegraphics[scale=0.8]{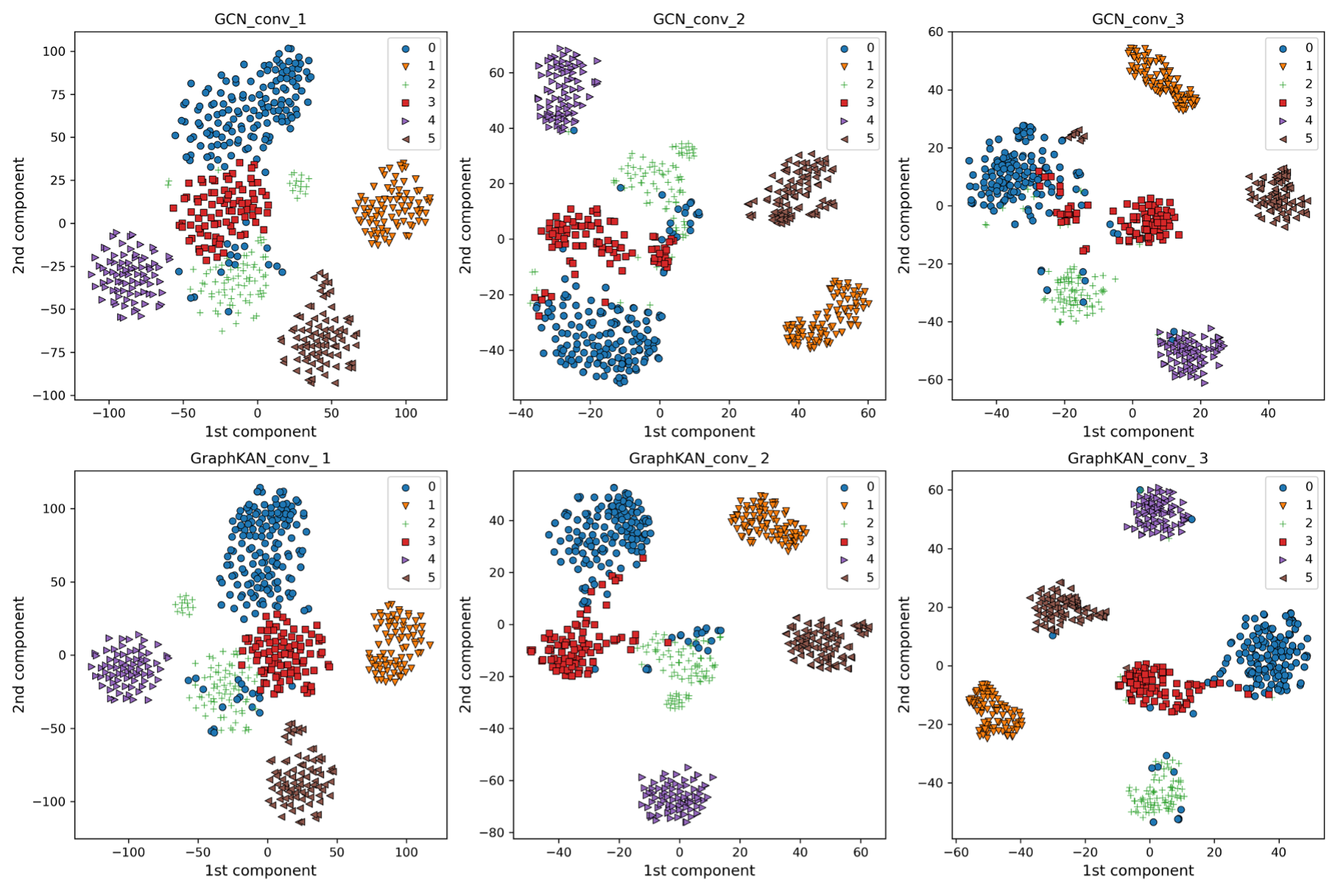}
\caption{Clustering comparison of intermediate features for testing nodes within BG\_2}
\label{fig3}
\end{figure}

\begin{figure}[!t]
\centering
\includegraphics[scale=0.8]{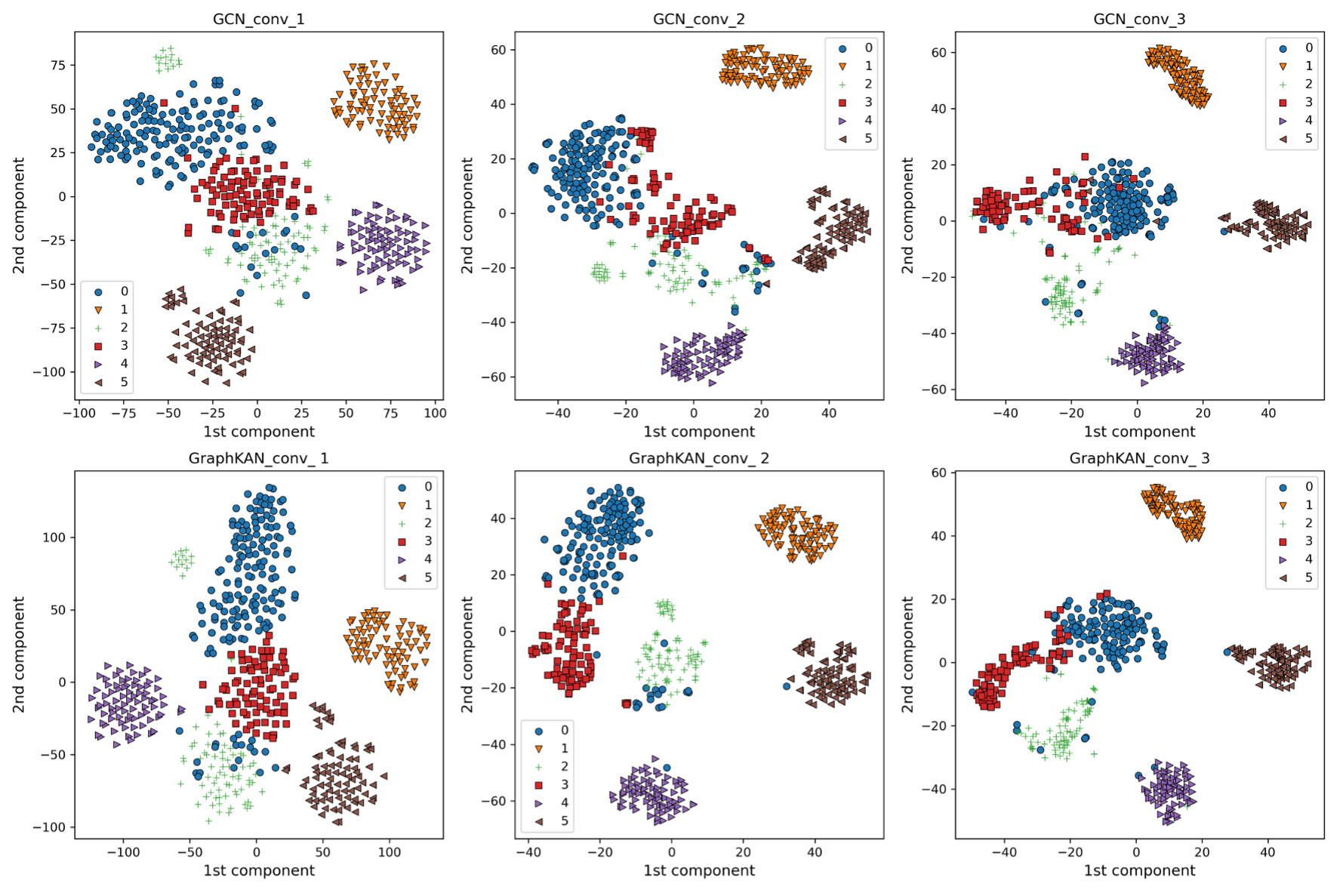}
\caption{Clustering comparison of intermediate features for testing nodes within BG\_3}
\label{fig4}
\end{figure}

\begin{figure}[!t]
\centering
\includegraphics[scale=0.8]{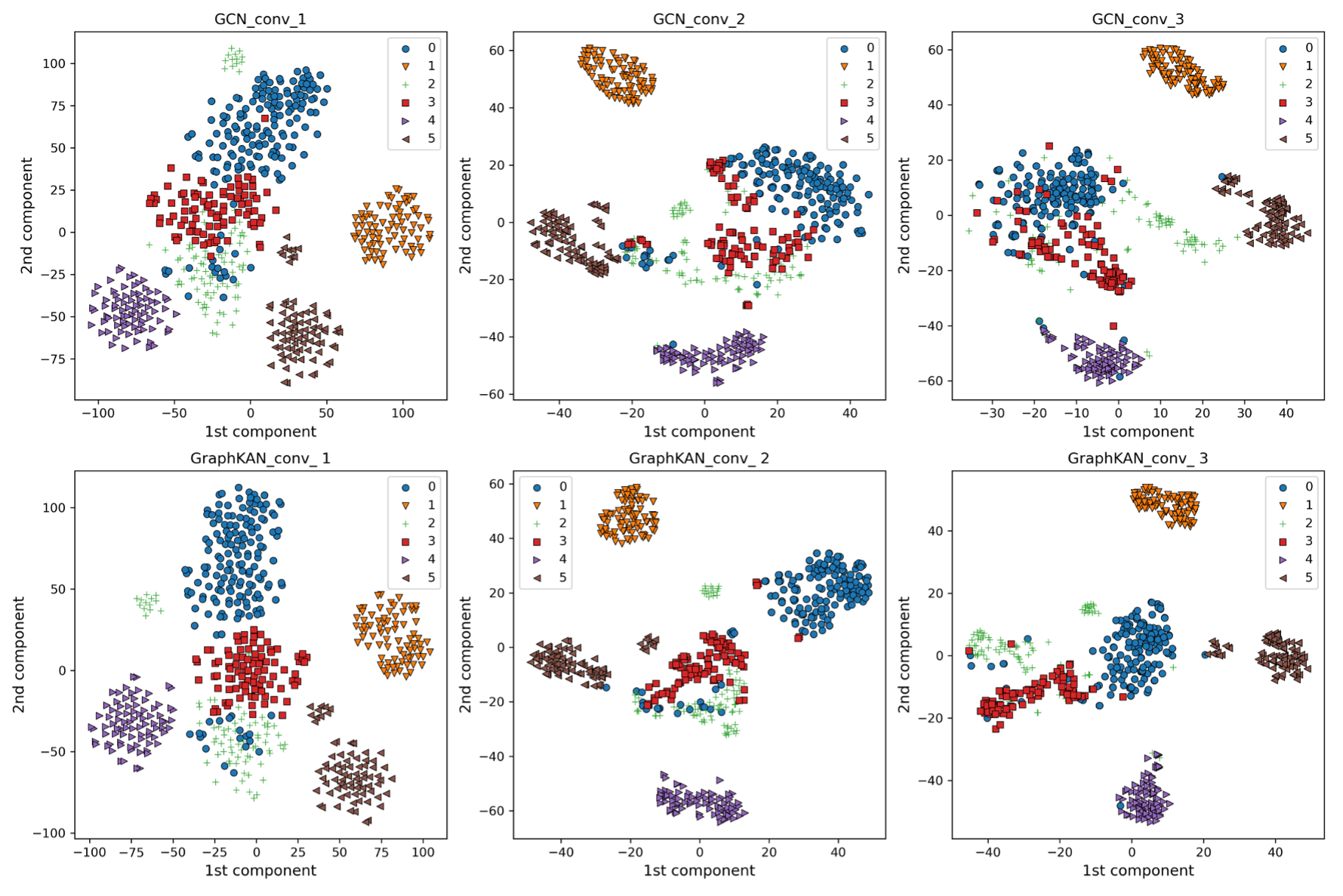}
\caption{Clustering comparison of intermediate features for testing nodes within BG\_3}
\label{fig5}
\end{figure}
A node classification task is constructed to validate the superiority of GraphKAN. It derives from a pattern recognition task for one-dimensional signals that collected from one axial-flow pump [3]. A specific graph construction method is designed in [1], demonstrating the relationship between samples. In this part, the constructed graphs named basic graphs (BGs) are used for node classification to validate the improvement effect of GraphKAN. According to the difference of the number of labeled samples, we constructed four graphs, as shown in Table. \ref{tab:settings}. The labeled nodes can be used for model training and model validation, and the testing nodes within these four graphs are the same. Two models, i.e., GraphKAN and GCN are separately conducted for processing the BGs. Their hyper-parameters are all set to the same value, as shown in Table 2. In this study, we adopted the CosineAnnealingLR learning rate scheduler in PyTorch to dynamically adjust the learning rate. We set the maximum number of epochs to 200 and constrained the minimum learning rate to 1e-4. This configuration aims to adjust the learning rate through cosine annealing to balance the model's convergence speed and performance during training. The above settings are valid for both models.\citet{fujimoto2018addressing}

We randomly select 20\% of labeled nodes for model validation, and the validation accuracy serves as a criterion to determine the model's convergence. 700 unlabeled nodes are utilized for model testing. Ten trials involving model training, validation, and testing were carried out, and the average test accuracy and time duration using various input graphs are illustrated in Fig. 1. Clearly, the time consumption of GraphKAN is notably higher than that of the original GCN, attributed to the more time-consuming computation of KAN. However, it can be observed that the test accuracy of GraphKAN is superior to that of GCN, particularly in BG\_3 and BG\_4. Although the time consumption has increased significantly, the improvement in accuracy is even more valuable. Moreover, the time consumption remains within a manageable range, less than one minute. At the same time, we also noticed that the improvement effect of GraphKAN is more significant when the input graph has less labeled nodes, such as BG\_3 and BG\_4. It probably indicates that GraphKAN holds potential significance for few-shot classification tasks.

We also used t-SNE algorithm to cluster the intermediate features of GNN models. Specifically, the output features of three graph convolutional layers for testing nodes are extracted and clustered in Fig. \ref{fig2}-\ref{fig5} , with different input graphs. It can be seen that in the clustering of outputs at the corresponding layers of these two models, GraphKAN always manages to cluster features of the same type more closely, while clusters of different types are farther apart. This indicates that GraphKAN's feature extraction capability surpasses that of GCN, thereby enhancing the classification accuracy.

\bibliographystyle{unsrtnat}
\bibliography{preference}

\begin{thebibliography}{24}
\providecommand{\natexlab}[1]{#1}
\providecommand{\url}[1]{\texttt{#1}}
\expandafter\ifx\csname urlstyle\endcsname\relax
  \providecommand{\doi}[1]{doi: #1}\else
  \providecommand{\doi}{doi: \begingroup \urlstyle{rm}\Url}\fi

\bibitem[O'shea and Nash(2015)]{o2015introduction}
Keiron O'shea and Ryan Nash.
\newblock An introduction to convolutional neural networks.
\newblock \emph{arXiv preprint arXiv:1511.08458}, 2015.

\bibitem[Deng et~al.(2009)Deng, Dong, Socher, Li, Li, and Fei-Fei]{deng2009imagenet}
Jia Deng, Wei Dong, Richard Socher, Li-Jia Li, Kai Li, and Li~Fei-Fei.
\newblock Imagenet: A large-scale hierarchical image database.
\newblock In \emph{2009 IEEE conference on computer vision and pattern recognition}, pages 248--255. Ieee, 2009.

\bibitem[Wang et~al.(2017)Wang, Jiang, Qian, Yang, Li, Zhang, Wang, and Tang]{wang2017residual}
Fei Wang, Mengqing Jiang, Chen Qian, Shuo Yang, Cheng Li, Honggang Zhang, Xiaogang Wang, and Xiaoou Tang.
\newblock Residual attention network for image classification.
\newblock In \emph{Proceedings of the IEEE conference on computer vision and pattern recognition}, pages 3156--3164, 2017.

\bibitem[Stahlberg(2020)]{stahlberg2020neural}
Felix Stahlberg.
\newblock Neural machine translation: A review.
\newblock \emph{Journal of Artificial Intelligence Research}, 69:\penalty0 343--418, 2020.

\bibitem[He et~al.(2017)He, Liao, Zhang, Nie, Hu, and Chua]{he2017neural}
Xiangnan He, Lizi Liao, Hanwang Zhang, Liqiang Nie, Xia Hu, and Tat-Seng Chua.
\newblock Neural collaborative filtering.
\newblock In \emph{Proceedings of the 26th international conference on world wide web}, pages 173--182, 2017.

\bibitem[Cui et~al.(2019)Cui, Henrickson, Ke, and Wang]{cui2019traffic}
Zhiyong Cui, Kristian Henrickson, Ruimin Ke, and Yinhai Wang.
\newblock Traffic graph convolutional recurrent neural network: A deep learning framework for network-scale traffic learning and forecasting.
\newblock \emph{IEEE Transactions on Intelligent Transportation Systems}, 21\penalty0 (11):\penalty0 4883--4894, 2019.

\bibitem[Fout et~al.(2017)Fout, Byrd, Shariat, and Ben-Hur]{fout2017protein}
Alex Fout, Jonathon Byrd, Basir Shariat, and Asa Ben-Hur.
\newblock Protein interface prediction using graph convolutional networks.
\newblock \emph{Advances in neural information processing systems}, 30, 2017.

\bibitem[Gligorijevi{\'c} et~al.(2021)Gligorijevi{\'c}, Renfrew, Kosciolek, Leman, Berenberg, Vatanen, Chandler, Taylor, Fisk, Vlamakis, et~al.]{gligorijevic2021structure}
Vladimir Gligorijevi{\'c}, P~Douglas Renfrew, Tomasz Kosciolek, Julia~Koehler Leman, Daniel Berenberg, Tommi Vatanen, Chris Chandler, Bryn~C Taylor, Ian~M Fisk, Hera Vlamakis, et~al.
\newblock Structure-based protein function prediction using graph convolutional networks.
\newblock \emph{Nature communications}, 12\penalty0 (1):\penalty0 3168, 2021.

\bibitem[Fan et~al.(2019)Fan, Ma, Li, He, Zhao, Tang, and Yin]{fan2019graph}
Wenqi Fan, Yao Ma, Qing Li, Yuan He, Eric Zhao, Jiliang Tang, and Dawei Yin.
\newblock Graph neural networks for social recommendation.
\newblock In \emph{The world wide web conference}, pages 417--426, 2019.

\bibitem[Zhou et~al.(2020)Zhou, Cui, Hu, Zhang, Yang, Liu, Wang, Li, and Sun]{zhou2020graph}
Jie Zhou, Ganqu Cui, Shengding Hu, Zhengyan Zhang, Cheng Yang, Zhiyuan Liu, Lifeng Wang, Changcheng Li, and Maosong Sun.
\newblock Graph neural networks: A review of methods and applications.
\newblock \emph{AI open}, 1:\penalty0 57--81, 2020.

\bibitem[Xu et~al.(2018)Xu, Hu, Leskovec, and Jegelka]{xu2018powerful}
Keyulu Xu, Weihua Hu, Jure Leskovec, and Stefanie Jegelka.
\newblock How powerful are graph neural networks?
\newblock \emph{arXiv preprint arXiv:1810.00826}, 2018.

\bibitem[Scarselli et~al.(2008)Scarselli, Gori, Tsoi, Hagenbuchner, and Monfardini]{scarselli2008graph}
Franco Scarselli, Marco Gori, Ah~Chung Tsoi, Markus Hagenbuchner, and Gabriele Monfardini.
\newblock The graph neural network model.
\newblock \emph{IEEE transactions on neural networks}, 20\penalty0 (1):\penalty0 61--80, 2008.

\bibitem[Vignac et~al.(2020)Vignac, Loukas, and Frossard]{vignac2020building}
Clement Vignac, Andreas Loukas, and Pascal Frossard.
\newblock Building powerful and equivariant graph neural networks with structural message-passing.
\newblock \emph{Advances in neural information processing systems}, 33:\penalty0 14143--14155, 2020.

\bibitem[Zhang et~al.(2019{\natexlab{a}})Zhang, Tong, Xu, and Maciejewski]{zhang2019graph}
Si~Zhang, Hanghang Tong, Jiejun Xu, and Ross Maciejewski.
\newblock Graph convolutional networks: a comprehensive review.
\newblock \emph{Computational Social Networks}, 6\penalty0 (1):\penalty0 1--23, 2019{\natexlab{a}}.

\bibitem[Veli{\v{c}}kovi{\'c} et~al.(2017)Veli{\v{c}}kovi{\'c}, Cucurull, Casanova, Romero, Lio, and Bengio]{velivckovic2017graph}
Petar Veli{\v{c}}kovi{\'c}, Guillem Cucurull, Arantxa Casanova, Adriana Romero, Pietro Lio, and Yoshua Bengio.
\newblock Graph attention networks.
\newblock \emph{arXiv preprint arXiv:1710.10903}, 2017.

\bibitem[Zhang et~al.(2019{\natexlab{b}})Zhang, Song, Huang, Swami, and Chawla]{zhang2019heterogeneous}
Chuxu Zhang, Dongjin Song, Chao Huang, Ananthram Swami, and Nitesh~V Chawla.
\newblock Heterogeneous graph neural network.
\newblock In \emph{Proceedings of the 25th ACM SIGKDD international conference on knowledge discovery \& data mining}, pages 793--803, 2019{\natexlab{b}}.

\bibitem[Taud and Mas(2018)]{taud2018multilayer}
Hind Taud and Jean-Franccois Mas.
\newblock Multilayer perceptron (mlp).
\newblock \emph{Geomatic approaches for modeling land change scenarios}, pages 451--455, 2018.

\bibitem[Friedland and Krell(2017)]{friedland2017capacity}
Gerald Friedland and Mario Krell.
\newblock A capacity scaling law for artificial neural networks.
\newblock \emph{arXiv preprint arXiv:1708.06019}, 2017.

\bibitem[Barcel{\'o} et~al.(2020)Barcel{\'o}, Monet, P{\'e}rez, and Subercaseaux]{barcelo2020model}
Pablo Barcel{\'o}, Mika{\"e}l Monet, Jorge P{\'e}rez, and Bernardo Subercaseaux.
\newblock Model interpretability through the lens of computational complexity.
\newblock \emph{Advances in neural information processing systems}, 33:\penalty0 15487--15498, 2020.

\bibitem[Yarotsky(2017)]{yarotsky2017error}
Dmitry Yarotsky.
\newblock Error bounds for approximations with deep relu networks.
\newblock \emph{Neural Networks}, 94:\penalty0 103--114, 2017.

\bibitem[Liu et~al.(2024)Liu, Wang, Vaidya, Ruehle, Halverson, Solja{\v{c}}i{\'c}, Hou, and Tegmark]{liu2024kan}
Ziming Liu, Yixuan Wang, Sachin Vaidya, Fabian Ruehle, James Halverson, Marin Solja{\v{c}}i{\'c}, Thomas~Y Hou, and Max Tegmark.
\newblock Kan: Kolmogorov-arnold networks.
\newblock \emph{arXiv preprint arXiv:2404.19756}, 2024.

\bibitem[Xu et~al.(2019)Xu, Sun, Zhang, Zhao, and Lin]{xu2019understanding}
Jingjing Xu, Xu~Sun, Zhiyuan Zhang, Guangxiang Zhao, and Junyang Lin.
\newblock Understanding and improving layer normalization.
\newblock \emph{Advances in neural information processing systems}, 32, 2019.

\bibitem[Zhang et~al.(2024{\natexlab{a}})Zhang, Jiang, Wang, Zhang, and Zhang]{zhang2024pruned}
Xin Zhang, Li~Jiang, Lei Wang, Tianao Zhang, and Fan Zhang.
\newblock A pruned-optimized weighted graph convolutional network for axial flow pump fault diagnosis with hydrophone signals.
\newblock \emph{Advanced Engineering Informatics}, 60:\penalty0 102365, 2024{\natexlab{a}}.

\bibitem[Zhang et~al.(2024{\natexlab{b}})Zhang, Liu, Zhang, and Lu]{zhang2024multiscale}
Xin Zhang, Jie Liu, Xi~Zhang, and Yanglong Lu.
\newblock Multiscale channel attention-driven graph dynamic fusion learning method for robust fault diagnosis.
\newblock \emph{IEEE Transactions on Industrial Informatics}, 2024{\natexlab{b}}.

\end{thebibliography}

\end{document}